\documentclass[pdflatex,sn-mathphys-num, iicol]{sn-jnl}% Math and Physical Sciences Numbered Reference Style
\usepackage{graphicx}%
\usepackage{multirow}%
\usepackage{amsmath,amssymb,amsfonts}%
\usepackage{amsthm}%
\usepackage{mathrsfs}%
\usepackage[title]{appendix}%
\usepackage{xcolor}%
\usepackage{textcomp}%
\usepackage{manyfoot}%
\usepackage{booktabs}%
\usepackage{algorithm}%
\usepackage{algorithmicx}%
\usepackage{algpseudocode}%
\usepackage{listings}%
\usepackage{caption}
\usepackage{flushend}

\theoremstyle{thmstyleone}%
\theoremstyle{thmstyletwo}%

\theoremstyle{thmstylethree}%
\flushend
\begin{document}

\title[Article Title]{Driving-RAG: Driving Scenarios Embedding, Search, and RAG Applications}


\author[1]{\fnm{Cheng} \sur{Chang}}%\email{changche21@mails.tsinghua.edu.cn}

\author[1]{\fnm{Jingwei} \sur{Ge}}%\email{gjw19@mails.tsinghua.edu.cn}

\author[2]{\fnm{Jiazhe} \sur{Guo}}%\email{gjz23@mails.tsinghua.edu.cn}

\author[1]{\fnm{Zelin} \sur{Guo}}%\email{gzl23@mails.tsinghua.edu.cn}

\author[1]{\fnm{Binghong} \sur{Jiang}}%\email{jbh23@mails.tsinghua.edu.cn}

\author*[1]{\fnm{Li} \sur{Li}}\email{li-li@tsinghua.edu.cn}

\affil[1]{\orgdiv{Department of Automation}, \orgname{Tsinghua University}, \orgaddress{\city{Beijing}, \postcode{100084}, \country{China}}}

\affil[2]{\orgdiv{Shenzhen International Graduate School}, \orgname{Tsinghua University}, \orgaddress{\city{Shenzhen}, \postcode{518055}, \country{China}}}


\abstract{Driving scenario data play an increasingly vital role in the development of intelligent vehicles and autonomous driving. Accurate and efficient scenario data search is critical for both online vehicle decision-making and planning, and offline scenario generation and simulations, as it allows for leveraging the scenario experiences to improve the overall performance. Especially with the application of large language models (LLMs) and Retrieval-Augmented-Generation (RAG) systems in autonomous driving, urgent requirements are put forward. In this paper, we introduce the Driving-RAG framework to address the challenges of efficient scenario data embedding, search, and applications for RAG systems. Our embedding model aligns fundamental scenario information and scenario distance metrics in the vector space. The typical scenario sampling method combined with hierarchical navigable small world can perform efficient scenario vector search to achieve high efficiency without sacrificing accuracy. In addition, the reorganization mechanism by graph knowledge enhances the relevance to the prompt scenarios and augment LLM generation. We demonstrate the effectiveness of the proposed framework on typical trajectory planning task for complex interactive scenarios such as ramps and intersections, showcasing its advantages for RAG applications.}




\maketitle

\section{Introduction}\label{sec1}
Driving scenarios are typically defined as comprehensive representations of the environment and driving behaviors within a specific temporal and spatial range \cite{chang2023metascenario, li2022features}. They systematically describe the states and tasks of various traffic participants, as well as the surrounding environment, including the road network and infrastructure. Scenario data have played an important role in tasks related to intelligent vehicles and robotics, such as prediction \cite{hu2022scenario, zhang2022systematic}, planning \cite{duan2024distributional, li2023onramp}, control \cite{hu2024noise, huang2024mfe, li2019integral}, and testing \cite{ge2024task, feng2021intelligent}.

In the era of artificial intelligence and large models, the developments of autonomous driving and intelligent vehicle systems increasingly rely on massive scenario data. On the one hand, scenario data need to be well stored and labeled to assist the efficient training and testing of models for autonomous driving. The typical scenario platforms include OpenScenario \cite{openscenario}, MetaScenario \cite{chang2023metascenario}, and CommonRoad \cite{althoff2017commonroad}, etc. In particular, previous works on scenarios enable efficient data storage and propose graph dynamic time warping (Graph-DTW) metric for labeling complex interactions and corner scenarios, facilitating the collection of large amounts of valuable samples \cite{chang2023metascenario} \cite{chang2024vista}.

\begin{figure*}[htbp]
\centering
% \captionsetup{justification=centering, labelfont={color=blue}}
\includegraphics[width=5.75in]{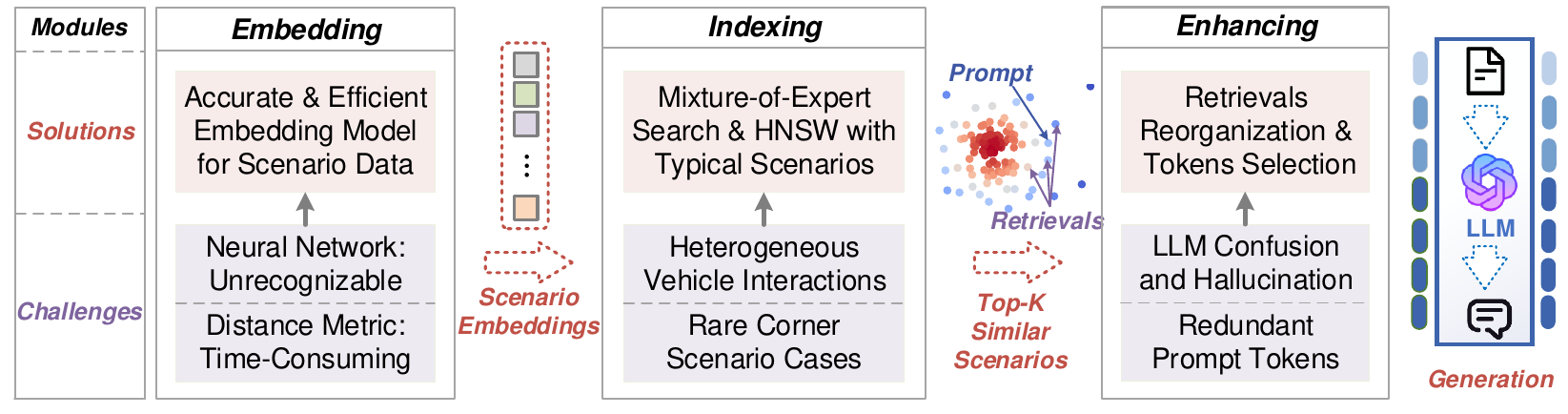}
\caption{The challenges and corresponding solutions in Driving-RAG framework.}
\label{fig_0}
\end{figure*}

On the other hand, efficient search capabilities are also essential for scenario data. Many researches have shown that leveraging previous experiences and knowledge enables vehicles to make more informed decisions and perform driving tasks effectively \cite{cai2024driving, wang2023chatgpt, chang2024llm}. A prominent application is the integration of Large Language Models (LLMs) with Retrieval-Augmented Generation (RAG) systems \cite{hussien2024rag, yuan2024rag}, which generate accurate and contextually relevant responses by providing timely access to similar scenarios. The RAG framework can support both online applications, such as vehicle planning \cite{xu2024drivegpt4} and decision-making \cite{dai2024tiv}, and offline applications, such as scenario generation \cite{nguyen2024text} and simulation \cite{guo2024mixing, zhang2023cavsim}. However, achieving efficient retrieval of similar scenarios and enhancing RAG applications remain challenging and require further development.

First, conventional scenario embedding approaches encounter the problems of accuracy or efficiency, and need to be aligned in vector space to better distinguish scenarios for efficient search. At present, mature pre-trained embedding models mostly focus on perception data. Unlike high-dimensional embeddings from complex image/video data, scenario data require less redundant encoding. Concise embeddings can effectively measure scenario differences, which reduce computational load and simplify retrieval from database. Although researchers have proposed some scenario embedding models via typical tasks, such as collision prediction \cite{malawade2022roadscene2vec}, risk assessment \cite{wang2024rs2g}, and scenario restoration \cite{wang2020clustering}, the embeddings mainly rely on the self-learning of the neural network. Sometimes, the embedded hidden representations are not so recognizable.

In addition, there exist some criteria that measure the distance between driving scenarios, such as slot \cite{kerber2020clustering}, dynamic time warping (DTW) \cite{hauer2020clustering}, contrastive distance \cite{zipfl2023traffic}, and Graph DTW \cite{chang2024vista} methods. The scenarios can be compared and then embedded into vector space according to the distances. However, the pair-to-pair computation style is more time consuming, and it is difficult for new scenarios to directly obtain the embeddings. A more accurate and efficient model that aligns the above features should be designed.

Second, an efficient multi-level scenario similarity search method specific to scenario distribution should be established. At the first level, various types of driving scenarios encompass heterogeneous vehicle behaviors and interactions \cite{hu2022scenario} \cite{chang2024vista}. Thus, instead of the unified search method, an accurate and efficient search can be achieved with a multi-expert indexing architecture, where each branch serves as an expert head for a particular interaction type, allowing retrieval of scenarios that most closely match the prompt type.

At the second level, even within the same expert indexing cluster, scenario data remain diverse. More important corner scenarios are mainly concentrated at the edge of the distribution space \cite{chang2024vista,li2021foundation}, which contain more valuable information that deals with difficult and complex cases. In contrast, the distances of most normal scenarios are relatively close and clustered, which causes the waste of storage and computation resources. Taking the hierarchical navigable small world (HNSW) method \cite{malkov2020approximate} as example, which currently has almost the best performance for vector search, the dense clusters will hinder the search process. When a prompt scenario represents a corner case, substantial time may be spent navigating out of dense clusters to locate sparse and relevant samples, potentially leading to premature termination without achieving the retrieval goal. Conversely, for a common scenario prompt, precision is less critical, as a subset of representative scenarios can effectively replace the nearest exact matches. The scenario search process should consider these issues.

Third, to enhance the RAG applications, the retrieval results need to be reorganized. Considering the relatively limited interpretability of vector embeddings, scenarios beyond the nearest one in the database may still rank among the closest matches and could serve as valuable references. Thus, it is beneficial to search for top-$K$ scenarios closest to the prompt. Within the small retrieval set, applying graph relations and knowledge extraction \cite{wang2024rs2g, pan2024unifying} from native scenario data can help reorganize the search results to better align with the prompt scenario. In addition, if the nearest searched scenario is significantly different from the prompt, the rare scenario will be added to the database for future reference. Inappropriate searched empirical cases should be discarded to prevent LLM confusion and hallucination. Moreover, the description tokens fed into LLM should be properly selected, not only to reduce the computation cost, but also to enhance LLM contextual understanding to capture critical information.

To solve the above issues, in this paper, we propose Driving-RAG framework to facilitate the scenarios embedding, indexing, and enhancing for RAG applications. As shown in Fig. \ref{fig_0}, we illustrate the challenges and corresponding solutions in the framework.

Our contributions are mainly listed as follows:

1) To address the unrecognizable embeddings by self-learning neural network methods, and the time-consuming issues by detailed scenario distance metrics, we propose an aligned scenario embedding model, where the training process combines the advantages of both types of methods. The model based on graph convolution and attention mechanism can effectively encode scenarios and achieve alignment in the feature space. The appropriate dimension for scenario embedding and the functions of each parts are also investigated in experiments.

2) With the aligned scenario embeddings, considering the heterogeneous vehicle interactions and behaviors, we propose a multi-level scenario search architecture with the classified interactive scenarios. In each expert interaction set, we design a novel vector similarity search method based on the scenario data distribution. Typical scenario data sampling combined with HNSW (\textit{HNSW-TSD}) can achieve more efficient indexing with almost consistent accuracy. We also compare the performance of other search methods, and investigate the influences caused by the scale of data.

3) With top-$K$ retrieved scenarios, considering the relatively limited interpretability of vector embeddings, we further perform reorganization by relations extraction and level selection to obtain the most relevant ones to avoid LLM's confusion and hallucination. The effectiveness of our scenario data infrastructure for augmenting typical LLM-RAG planning task is verified on different types of scenarios in the \textit{CitySim} and \textit{INTERACTION} datasets.

The rest of this paper is arranged as follows. \textit{Section 2} introduces the embedding, search, and RAG enhancing framework specific to scenario data. \textit{Section 3} verifies the effectiveness of the proposed framework via comparison experiments. Finally, \textit{Section 4} concludes the paper.

\section{The Driving-RAG Framework}
In this section, we introduce the Driving-RAG framework in detail, which contains the aligned scenario embedding model, the \textit{HNSW-TSD} algorithm for scenario vector similarity search, and retrieval reorganization for augmenting LLM generation.

\begin{figure*}[htbp]
\centering
\includegraphics[width=6in]{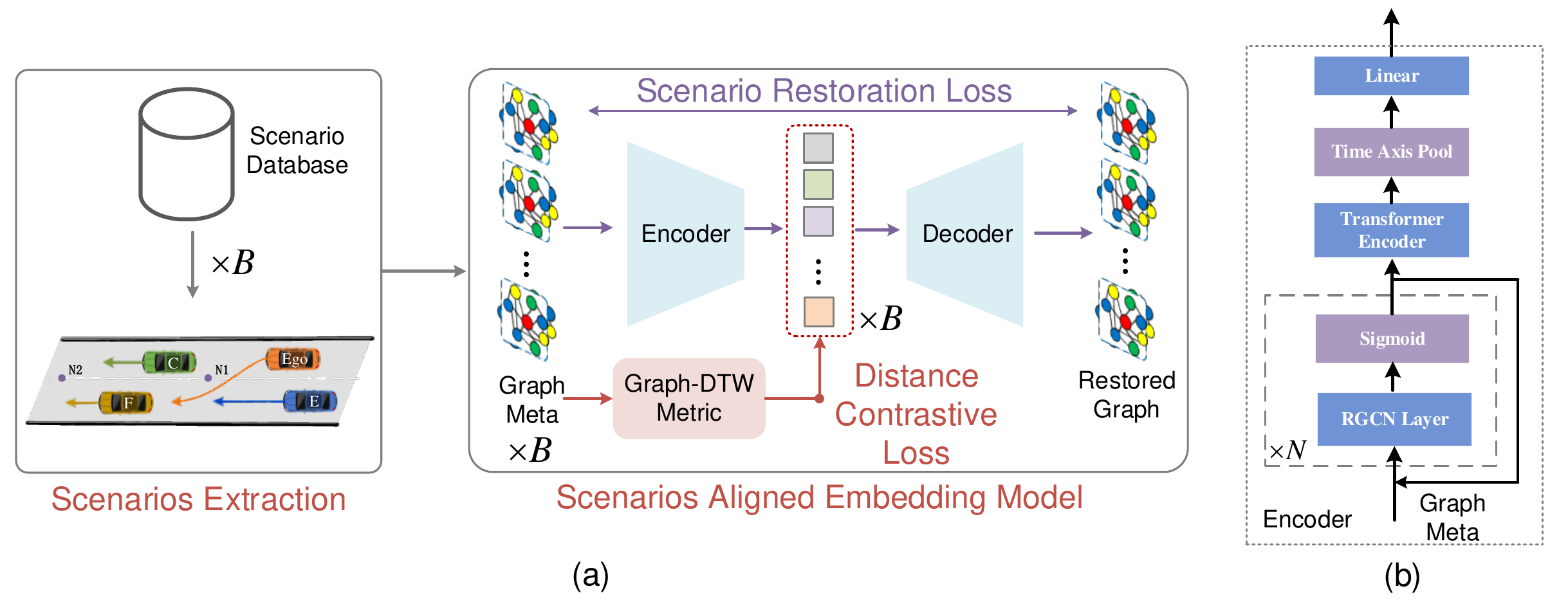}
\caption{(a) The proposed scenario embedding model and training process. (b) The backbone of the scenario graphs encoder.}
\label{fig_1}
\end{figure*}

\subsection{Aligned Scenario Embedding Model}
Based on previous works \cite{chang2023metascenario, chang2024vista}, with scenario data stream, scenario segments (i.e., atom scenarios) are sliced and classified according to different vehicle interaction types. Each atom scenario contains independent and complete decision-making process within a spatial and temporal range. Furthermore, each scene/frame of the atom scenario is represented and stored as a graph structure, which has the advantages of rich semantic relations and state descriptions of traffic participants and can be easily converted to text tokens for LLM \cite{chang2024llm}. The following “scenarios” in this paper all refer to the sliced scenario segments.

To obtain accurate scenario vector embeddings in an efficient way, we train an aligned scenario embedding model through graph restoration and scenario distance fitting tasks.

The backbone embedding neural network is shown in Fig. \ref{fig_1}. Time series of graph data are sequentially embedded by Relational Graph Convolutional Network (RGCN) and Transformer. In ${(i+1)}^{th}$ layer of RGCN, the hidden feature of graph node $i$ is represented as:

\begin{equation}
\label{equ1}
f_i^{l+1} =\sigma\left(\sum_{r \in R} \sum_{j \in N^r_i} \frac{1}{c_i^r d_{i, j}} W_r^l f_j^l+W_0^l f_i^l\right)
\end{equation}

\begin{equation}
\begin{aligned}
R &= \{ \text{front}, \text{front\_left}, \text{front\_right}, \\
  &\quad \quad \text{rear}, \text{rear\_left}, \text{rear\_right} \}
\end{aligned}
\end{equation}

\begin{equation}
c_i^r=\alpha\left|N_i^r\right|, \quad \alpha_{V2V} / \alpha_{V2N}=1 / 4
\end{equation}
where relations $R$ is the set of relative directions between surrounding vehicles and the ego \cite{chang2023metascenario}, $N_i^r$ denotes the set of neighbor indices of node $i$ under relation $r \in R, W_r^l$ is learnable weight that is shared by all edges of relation $r$ in layer $l$, $\sigma$ is the sigmoid activation function, $c_i^r$ is normalized variable considering both the type and number of graph relations, $\alpha_{V2V} / \alpha_{V2N}$ indicates that the message passed by vehicle-to-vehicle relations is weighted more than that passed by vehicle-to-roadnode relations, and $d_{i, j}$ is the node $i, j$'s distance that indicates the interaction strength.

The features output by RGCN will summarize, and then capture connections between temporal graph contexts via multi-head attention mechanism. The number of attention heads is $h$, the embedding dimension is $d$, and the projections of each attention head are $Q_i \in R^{d_q}, K_i \in R^{d_i}, V_i \in R^{d_i}$. The parameters include $h$  groups of weight matrix pairs $\left(W_i^q, W_i^k, W_i^v\right)$ corresponding to dimension $\left(d \times d_q, d \times d_k, d \times d_v\right)$.
Thus the query, key and value are:

\begin{equation}
\label{equ2}
Q_i=q W_i^q, K_i=k W_i^k, V_i=v W_i^v
\end{equation}

The attention tensor of each head is:
\begin{equation}
\label{equ3}
\textit{head}_i=\operatorname{Att}\left(Q_i, K_i, V_i\right)=\operatorname{softmax}\left(\frac{Q_i K_i^T}{\sqrt{d_k}}\right) V_i
\end{equation}

By concatenating multi-head results, the final attention is as follows:
\begin{equation}
\label{equ4}
S=\textit{Concat}\left(\textit{head}_1, \ldots, \textit{head}_h\right) W^0
\end{equation}
where the $W^0$ dimension is $d \times d, S$ is the potential scenario embedding.

To train the model, first, we employ RGCN features within an autoencoder architecture for the graph restoration task, which predicts the connection relationships of graphs. While the self-learning capabilities of the network allow for the extraction of basic features, they fall short of capturing more recognizable and complex features necessary for effective scenario comparison.

Second, we utilize the Graph-DTW scenario distance metric shown in Fig. \ref{fig_2}, which integrates optimal transport and DTW to calculate scenario distances. Previous works \cite{chang2024vista, chang2024llm} have demonstrated that Graph-DTW effectively measures the differences between scenarios. While it provides accurate labeling of collected scenarios, the embedding task becomes time-consuming. By leveraging the distance labels from the training scenario set, the understanding and fitting capabilities of RGCN and Transformer models enable us to extract richer features that account for both semantic relationships and the spatio-temporal evolution of scenarios, ultimately leading to improved embeddings.

The training process is similar to contrast learning. On the one hand, basic scenario understanding is maintained through graph structure restoration and the constraints can prevent scenario distances overfitting. On the other hand, the extraction of spatio-temporal features brings the scenario embedding distances closer to the Graph-DTW distances where scenarios can be effectively distinguished. From the perspective of embedding performance, when the distances between the encoded embeddings align closely with those of the scenario metric, and when the restored graph exhibits a high Intersection over Union (IoU), we recognize that the model successfully extracts useful features and achieves alignment. The loss function of the training process is designed as:

\begin{figure*}[htbp]
\centering
\includegraphics[width=6.4in]{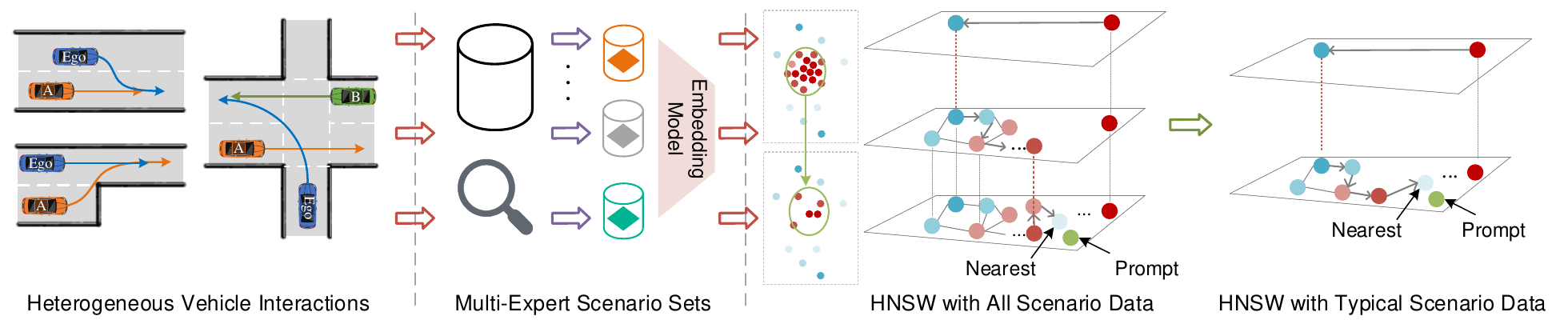}
\caption{The driving scenario embeddings similarity search process with multi-interactions and \textit{HNSW-TSD} algorithm.}
\label{fig_2}
\end{figure*}

\begin{algorithm*}
\caption{Indexing Module: HNSW-TSD}
\textbf{Input:} $S = \{s_1, s_2, \dots, s_n\}$ : Original scenario data, density quantile $\alpha\%$, sampling proportion $\beta\%$, distance threshold $D$, $S^p = \{ s^p_1, s^p_2, \dots, s^p_n\}$: $n$ prompt scenarios.

\textbf{Output:} $\textit{TSD}$: Typical scenario data; ${S}^{\textit{HNSW}}$ : Searched top-$K$ similar scenarios from HNSW.

\textbf{Steps:}
1. \textbf{Density Calculation}:
   \begin{itemize}
       \item Each scenario $s_i$ in $S$ database is embedded as a vector $\mathbf{v}_i \in \mathbb{R}^d$, compute the density $p(s_i)$ using Kernel Density Estimation (KDE) with appropriate bandwidth $W$.
   \end{itemize}

2. \textbf{Typical Scenario Selection}:
   \begin{itemize}
       \item \textbf{Low-Density Scenarios}: Retain all scenarios where the density $p(s_i)$ is less than the $\alpha\%$ quantile of the density distribution: $p(s_i) \leq \text{Quantile}_{\alpha\%}(p(S))$
       
       \item \textbf{High-Density Scenarios}: For scenarios where $p(s_i) > \text{Quantile}_{\alpha\%}(p(S))$, sample them with probability inversely proportional to their density: $P(\textit{select}(s_i)) = \frac{1}{p(s_i)} $
       with $\beta$\%:
       \[
       S_{\textit{high}} = \{s_i \in \text{$\beta$\% of } \{p(s_i) > \text{Quantile}_{\alpha\%}(p(S))\}\}
       \]

       \item Construct the typical scenario data set $\textit{TSD}$ as: $\textit{TSD} = S_{\textit{low}} \cup S_{\textit{high}}$
   \end{itemize}

3. \textbf{HNSW Construction}:
   \begin{itemize}
       \item Build the HNSW graph $\mathcal{G}^{\text{HNSW}}$ from $\textit{TSD}$. The edges are added between close vectors to form a hierarchical structure: 
       $
       \mathcal{G}^{\text{HNSW}} = \{ (\mathbf{v}_i, \mathbf{v}_j) \mid |N(\mathbf{v}_i)|, |N(\mathbf{v}_j)|\leq M \}
       $
       , where $M$ is the maximum neighbors for adding edges.
   \end{itemize}

4. \textbf{Scenario Retrieval}:
   \begin{itemize}
       \item Given prompt scenarios $S^p$, each scenario is embedded as $\mathbf{v}_i^p \in \mathbb{R}^d$. Perform nearest-neighbor search in $\mathcal{G}_{\text{HNSW}}$ to retrieve the top-$K$ most similar scenarios:
       $
       {S}_i^{\text{HNSW}} = \text{Top-$K$}(\{\mathbf{v}_j \mid \mathbf{v}_j \in \mathcal{G}^{\text{HNSW}}\}, \mathbf{v}_i^p)
       $
   \end{itemize}

5. \textbf{Scenario Expansion}:
   \begin{itemize}
       \item If $\min_{\mathbf{v}_j \in {S}_i^{\text{HNSW}}} d(\mathbf{v}^p_i, \mathbf{v}_j) > D$, it indicates the scenarios in database are inappropriate for the prompt. The scenario data corresponding to $v^p_i$ should be added to the database for collection and future reference.
   \end{itemize}
\label{alg}
\end{algorithm*}

\begin{equation}
\label{equ5}
\begin{aligned}
\textit{L}= \frac{1}{B} \sum_{i=1}^B \left(1 - \frac{1}{N_i} \sum_{j=1}^{N_i} \frac{|E^{\text{pred}}_{ij} \cap E_{ij}^l|} {|E^{\text{pred}}_{ij} \cup E^l_{ij}|} \right) +\\
\frac{2}{B(B-1)} \sum_{1\leq x < y \leq B} \left(d\left(S_x, S_y\right)-d_{x y}^l\right)^2 \\
\end{aligned}
\end{equation}
where $B$ is the batchsize, $N_i$ is the scene/frame numbers of $i^{\text {th}}$ scenario, $E^{\text{pred}}_{ij}$ and $E^l_{ij}$ indicate the predicted edge connections and labels for IoU calculation of adjacent edge matrix in the $j^{\text{th}}$ scene graph of $i^{\text {th}}$ scenario, and $d\left(S_x, S_y\right)$ and $d^l_{xy}$ are predicted scenario distance of $x^{\text{th}}$ and $y^{\text{th}}$ scenario and corresponding contrastive Graph-DTW distance.

In \textit{Section 3.2}, we will analyze the effectiveness of each part of the model, and investigate how many dimensions of embeddings can generally be enough to encode and distinguish scenarios well.

\subsection{HNSW-TSD Algorithm for Efficient Scenario Search}

Given the large number of scenarios, distinct interaction types have been classified and attributed to several expert sets based on vehicle motion flows, as outlined in previous works \cite{chang2024vista}. When vehicles engage in specific online or simulation tasks and seek assistance from the expert database, it is essential that the retrieved scenarios share the same interaction types as the prompt scenarios. The approach ensures that the searched results provide more relevant and accurate references.

Then, using the embedding model above, we can obtain scenario vectors and perform efficient search within database systems. As shown in Fig. \ref{fig_2}, dense clusters of common scenarios cause large space occupancy and waste of search resources. To this end, we design a new vector search mechanism \textit{HNSW-TSD} in Algorithm \ref{alg}, which combines with HNSW and typical scenario data (TSD). 
 
The algorithm supports flexible parameter adjustment based on the constructed scenario database. The density quantile, $\alpha\%$, should be set higher (80\%-90\%), while the sampling proportion for normal scenarios should be lower (3\%-10\%) to preserve sparse scenarios and a small subset of normal ones. The distance threshold $D$ should be set larger than the normal scenario distances to collect rare scenarios, the $K$ should be selected according to scenario complexity and task requirements, and the KDE bandwidth should be adjusted according to the data scale to ensure efficient computation. The first three steps only need to be executed once with long-cycle updates, and the last two steps conduct efficient search with frequent batch requests. This approach allows us to search for similar scenarios in a more efficient and accuracy-neutral way.

It is noted that there exist other vector similarity search methods \cite{douze2024faiss}, such as Flat search, Inverted Flat (IVF) \cite{babenko2014inverted}, and ProductQuantizer(PQ) \cite{xu2018online}. Our algorithm can effectively improve the search efficiency with the above base algorithms. In \textit{Section 3.3}, we will discuss the search/add performance of different search methods, data dimensions, and scales.

\subsection{Reorganize Scenarios for RAG Applications}
\begin{algorithm}
\caption{Enhancing Module: Retrieval Scenarios with Reorganization}
\textbf{Input}: $S^p = \{ s^p_1, s^p_2, \dots, s^p_n\}$: $n$ prompt scenarios for each task; $K$: Number of top similar scenarios for each $s^p_i$; $M$: Desired number of final retrieved scenarios for $S^p$.

\textbf{Output}:
$T_{\textit{RAG}}$: Scenarios for RAG applications after retrieval and reorganization.

\textbf{Steps}:

1. \textbf{Top-K Similar Scenarios Arrangements:}
\begin{itemize}
       \item Categorize all top-$K$ scenarios for $n$ prompts into $K$ levels as \(T = \{ T^1, T^2, \dots, T^K \}\), where
       \[
       T^l = \{ t_i^l \mid t_i^l \text{ is in the } l\text{-th level similarity for } s^p_i \}
       \]
\end{itemize}

2. \textbf{Reorganization with Graph Relations:}
\begin{itemize}
       \item For each $t_i^l \in T$, extract graph relations (e.g., vehicle-to-vehicle relations, vehicle-to-lane relations \cite{chang2023metascenario, chang2024llm}) in $s^p_i$ and $t_i^l$.
       
       \item If $t_i^l$ does not match $s^p_i$ relations, remove $t_i^l$ from $T^l$.
\end{itemize}
    
3. \textbf{Sequential Level Selection:}
\begin{itemize}
       \item For $l = 1, 2, \dots, K$, select scenarios from each level of $T$ in order as needed.
       \item If desired $M$ scenarios fulfill or all $K$ levels exhaust, break selection loop and obtain:
       \[
       T_{\textit{RAG}} = \bigcup_{l=1}^K T^{l'} \text{where } T^{l'} \subset T^l \text{ and } |T_{\textit{RAG}}| \leq M
       \]
\end{itemize}
\label{alg2}
\end{algorithm}

To use retrieval scenarios for enhancing LLM-RAG systems, we take trajectory planning task as example, which is a common RAG application \cite{wang2023chatgpt, yuan2024rag}. First, $n$ (usually 3-6) candidate trajectories are generated using $5^{th}$-degree polynomial in different future horizons for each vehicle task to consider multiple modals of vehicle behaviors and multiple possibilities for the scenarios \cite{zhang2022systematic}. Combined with the predicted states information of surrounding interacted vehicles, several potential scenarios are constructed and used for searching multiple similar scenarios. By judging the interaction type, we adopt corresponding embedding model to obtain scenario vectors, and use \textit{HNSW-TSD} algorithm to perform top-$K$ search in collected scenario database.

Then, for $n\times K$ retrieval scenarios, we conduct reorganization with graph relations extraction and level selection, as shown in Algorithm \ref{alg2}. While the scenario graph data are embedded as vectors to serve fast retrieval, the representation may still lack a certain degree of interpretability. Therefore, the native knowledge of graph relations, such as direction relations between vehicles and vehicle-lane connections \cite{chang2023metascenario, chang2024llm}, can be used for processing the small part of search results. Through the fusion of vector search and rule knowledge, it is guaranteed that the external scenario data used for RAG is relevant. Further by sequential level selection, $M$ ($M << nK$) scenarios are expected to conduct in-context learning in LLM for better performance \cite{wang2023chatgpt, barmann2024incremental}.

\begin{figure}[htbp]
\centering
\includegraphics[width=2.4in]{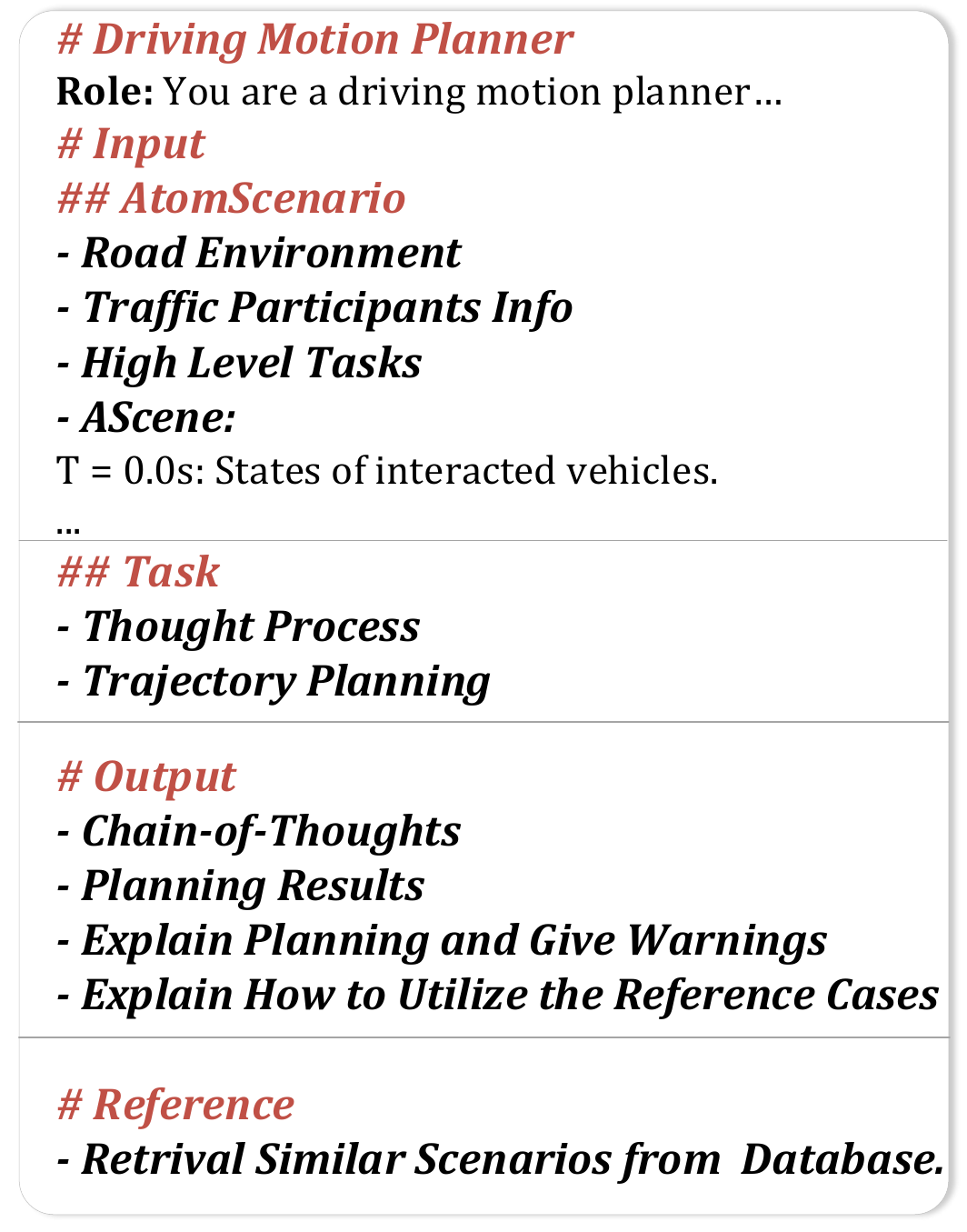}
\caption{The prompt texts for LLM for typical planning task.}
\label{fig_3}
\end{figure}

\begin{figure}[htbp]
\centering
\includegraphics[width=2.4in]{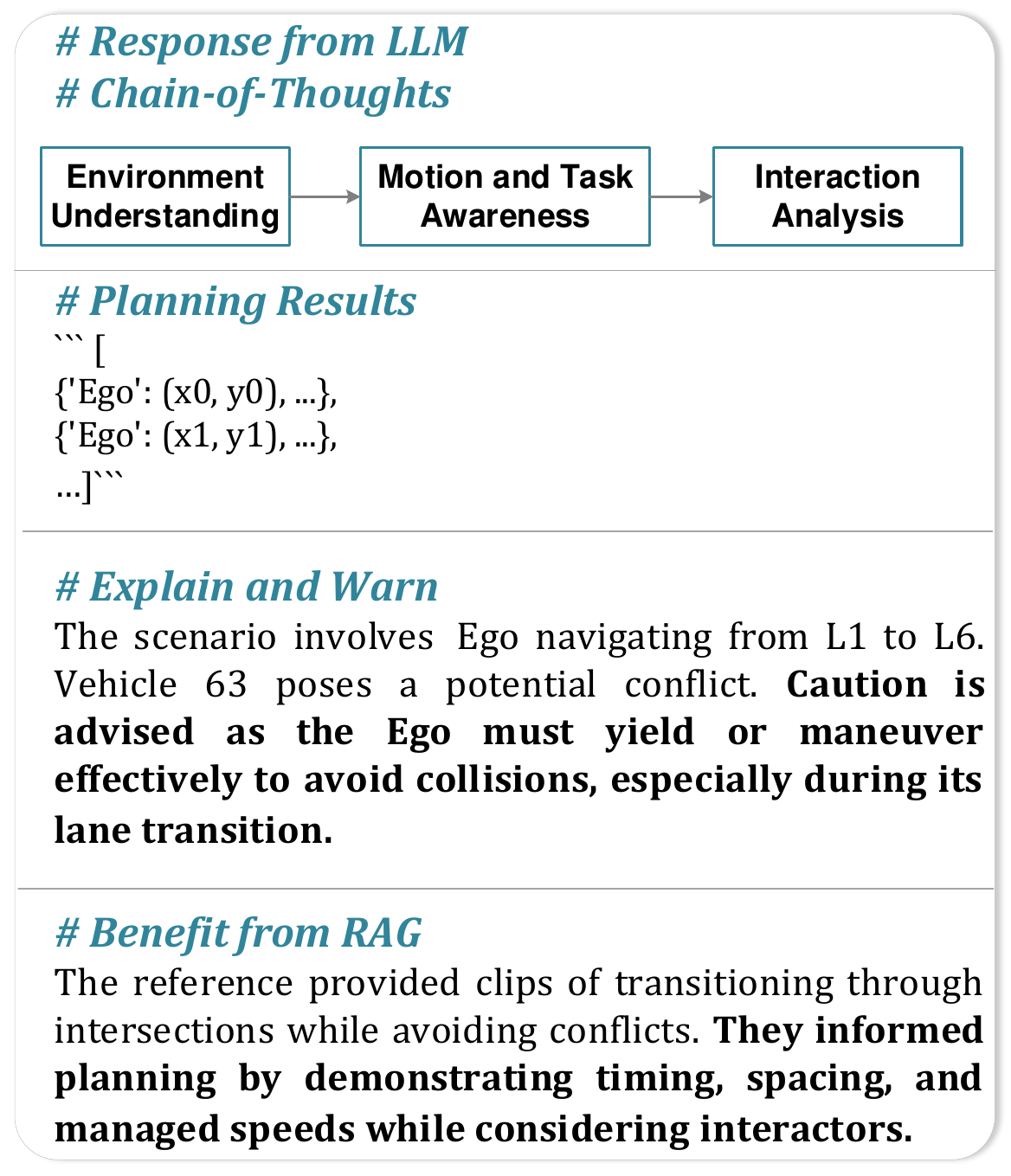}
\caption{The response texts from LLM for typical planning task.}
\label{fig_4}
\end{figure}

Finally, as shown in Fig. \ref{fig_3} \ref{fig_4}, we design the concise prompt with the information of instructions, scenarios, tasks, Chain-of-Thoughts (CoT), and searched reference cases based on previous works \cite{chang2024llm, guo2024mixing}. The LLM responses with the undertanding of CoT questions and provides the planning results. LLM further self-interprets the scenarios and provides appropriate warnings, and ensures that relevant knowledge is learned from external database.

\section{Experimental validation}
\subsection{Dataset and Settings}

\textbf{\textit{Datasets:}} 1) \textit{CitySim} \cite{zheng2024citysim}: \textit{CitySim} provides a large amount of vehicle trajectory data with high-precision aerial views by drones. One of the key features of \textit{CitySim} is the critical safety events. In particular, the FreeWayC section provides rich cut-in and on-ramp merging scenarios, which are suitable for studying vehicle trajectory data in complex traffic scenarios.

2) \textit{INTERACTION} \cite{zhan2019interaction}: \textit{INTERACTION} dataset is an international, adversarial, and cooperative motion dataset designed for interactive driving scenarios with semantic maps. It provides a rich set of data that can be used to evaluate methods for various aspects of vehicle interactions and traffic management. In particular, the Intersection-EP section provides busy intersection scenarios, which are suitable for studying vehicle motion behaviors in interactive situations.

\textbf{\textit{Settings:}} In our experiment, the scenarios are preliminary sliced, classified, and labeled in MetaScenario database \cite{chang2023metascenario, chang2024vista}. In this paper, we mainly focus on the complex interactive scenario types, i.e. conflict line in \textit{CitySim} FreewayC and conflict line\&point in \textit{INTERACTION} Intersection-EP. With the scenario database, we conduct the vector embedding, search, and RAG applications augmentation. The machine is equipped with an Intel 10900X CPU and two RTX 3090 devices. The operating system is Ubuntu 18.04LTS with 128G RAM.

\subsection{Validation of Embedding Model and Dimensions}
\begin{table*}[htbp]
\centering
% \captionsetup{justification=centering, labelfont={color=blue}}
\caption{Comparison of different embedding dimensions in scenario datasets.}
\includegraphics[width=6.4in]{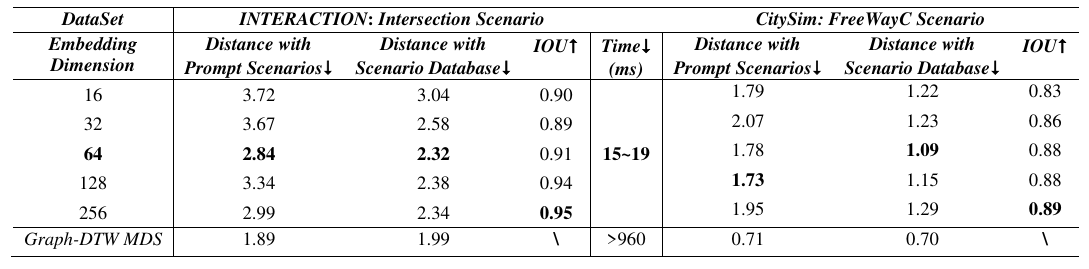}
\label{tab_1}
\end{table*}

The models in \textit{Section 2.1} embed prompt scenarios into vectors, which have been trained on thousands of graphs and tens of thousands of scanario distances. Here we validate the effectiveness of the model and investigate which number of embedding dimensions is better. As shown in Table \ref{tab_1}:

\begin{figure}[htbp]
\centering
\includegraphics[width=3.2in]{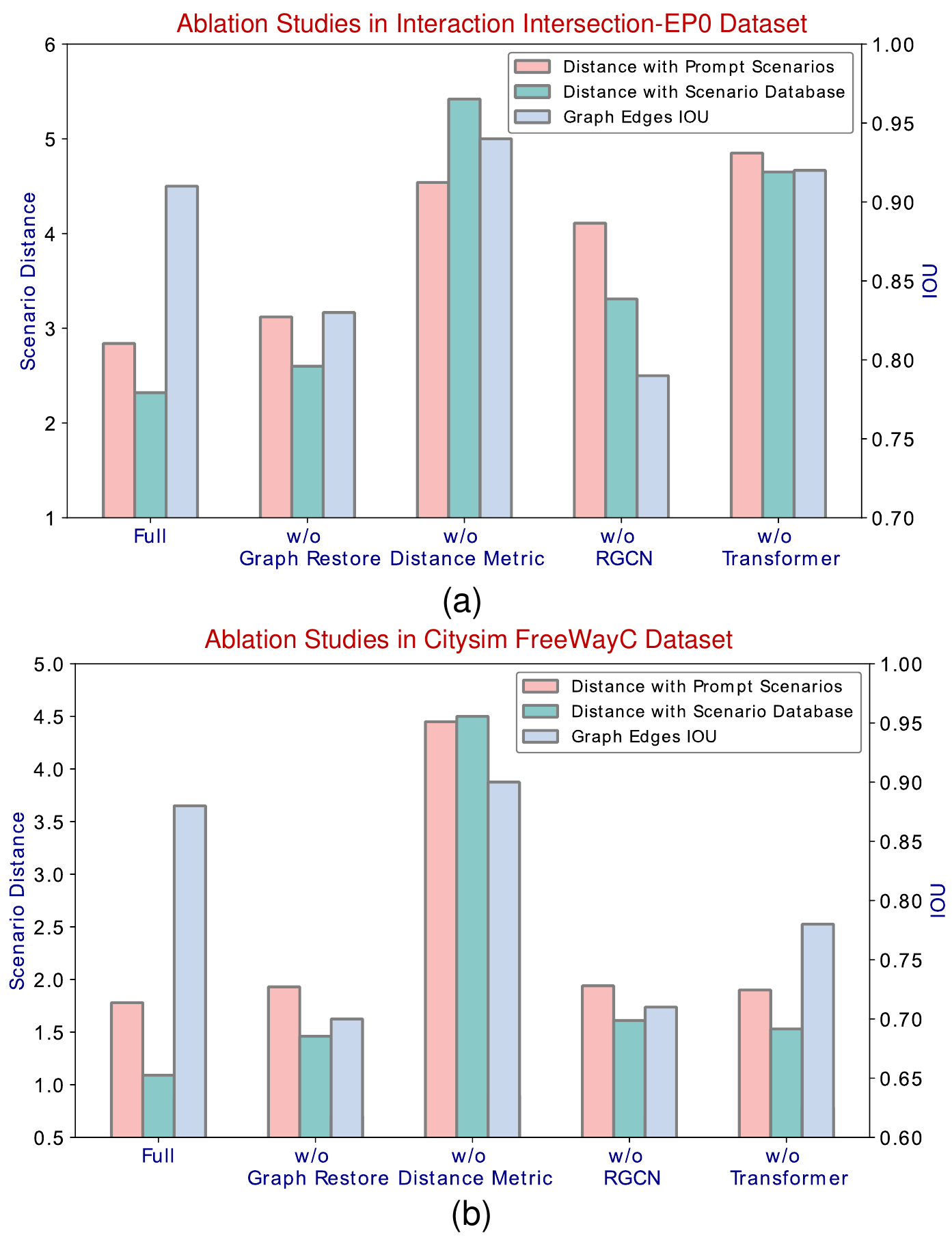}
\caption{Ablation studies for each part of embedding model in the two datasets.}
\label{fig_5}
\end{figure}

1) For both scenario types, the embeddings are close to the labeled distances by Graph-DTW (the distance is normalized in the vector space with the farthest being 100). It indicates that the model can measure the scenario differences, whether for the distances between the prompts and the database or between the other prompts. The IOU metric of the scenario graph structures is also high, which indicates that the model maintains fundamental scenario understanding and achieves features alignment.

2) When the embedding dimension is selected as 64 or 128, the embeddings are closest to the scenario distances. Although the IOU for graph structure restoration is promoted as the dimension increases, the distance bias increase. To further balance the search efficiency and accuracy, in this paper, we select 64 as the scenario embedding dimension.

3) Compared to the embeddings derived from the mathematical approach of Multi-Dimensional Scaling (MDS) with Graph-DTW, the distances derived by the deep neural network show a slight deviation. However, due to the high computational demands of MDS with Graph-DTW, which requires pairwise comparisons of scenarios and takes over 960 ms. In contrast, the neural network inference time for generating embeddings is only 15-19 ms, making it more efficient for vehicle tasks.

\begin{figure}[htbp]
\centering
\includegraphics[width=2.8in]{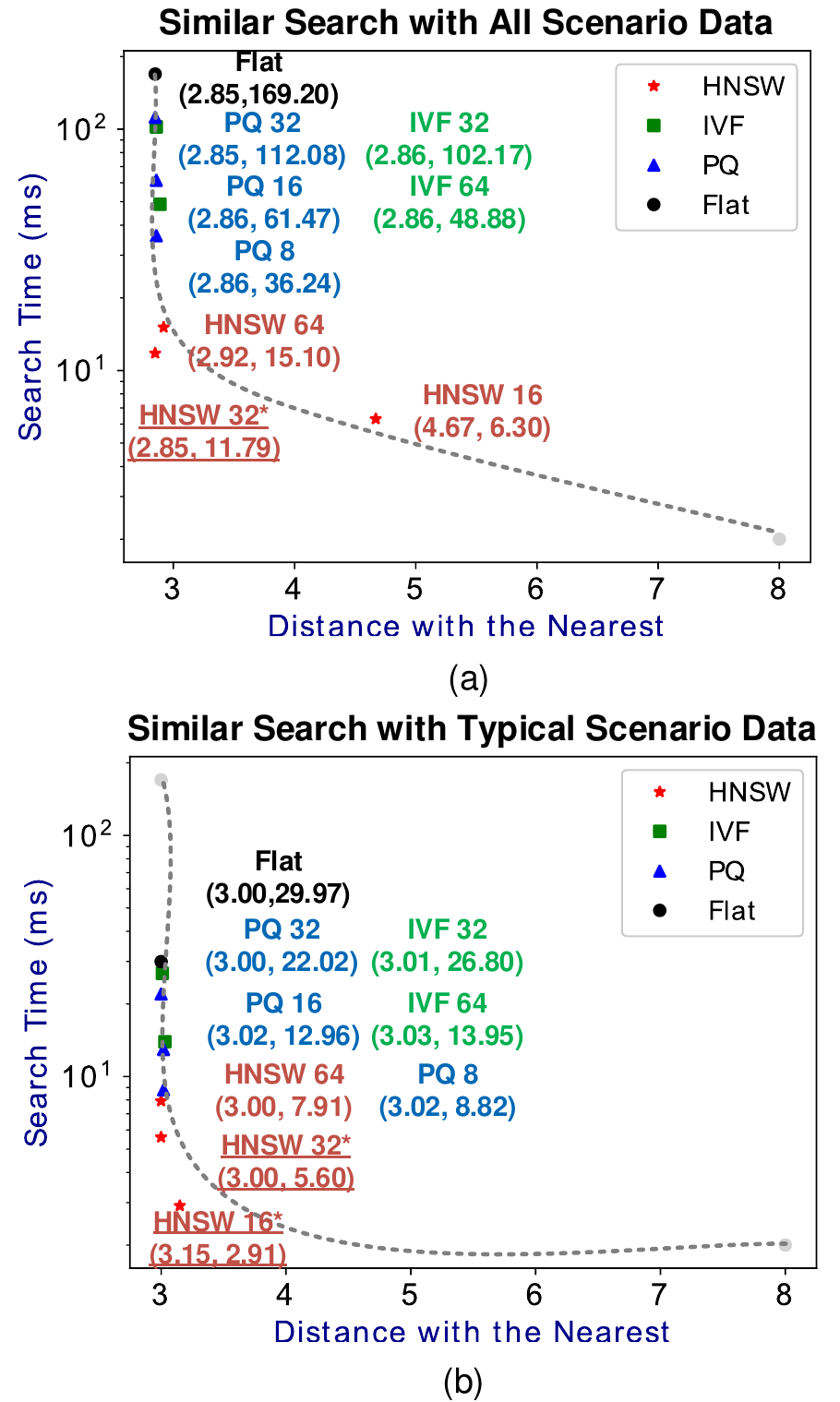}
\caption{Comparison of similarity search time with (a) All. (b)Typical data under different base algorithms (the y-axis is log-scale).}
\label{fig_6}
\end{figure}

Furthermore, we conduct ablation studies for each part of our designed model, as shown in Fig. \ref{fig_5}. First, without the aid of RGCN's parsing graphs or graph restoration task, the network's understanding of the scenario graphs becomes weaker, which is reflected in a sharp decrease in IOU, and also influences the learning of scenario distances. Second, without the aid of distance contrastive learning, the distance estimations will significantly deviate, which indicates the self-learning cannot learn so recognizable features to distinguish the scenarios. Third, the attention mechanism of Transformer can further promote both the learning of the two training tasks.

\subsection{Validation of HNSW-TSD Algorithm for Similarity Search}
In this part, we validate the \textit{HNSW-TSD} algorithm on a relatively larger data scale to better demonstrate its effectiveness. Since the amount of scenarios in existing datasets is not sufficient to simulate, we expand the embeddings to $10^4$ and $10^5$ levels of scale in the vector space by sampling and interpolating based on the density of scenarios.

First, in our simulation environment, we assume that the scenario cloud database has accumulated over $2\times 10^4$ scenarios, about 100 vehicles simultaneously request RAG's aid from the database, and each vehicle prompts n=5 scenarios for top-$4$ search with final M=4 desired in-context scenarios. The density quantile is $\alpha\%=90\%$, sampling portion is $\beta\%=5\%$, and $D=10$ for Algorithm \ref{alg}, and the embedding dimension is 64 in this experiment. The database conducts vector nearest-neighbor search in batch-style.

Our designed indexing mechanism are compared under different base searching methods, such as Flat, IVF, PQ. and HNSW. The brief introductions are listed as follows:

\textbf{Flat}: Search the nearest neighbor directly in vector dataset.

\textbf{IVF (Inverted File Index)}: IVF partitions the vector space into clusters and stores an inverted index of which vectors belong to each cluster. When searching, it first finds the closest cluster, and then searches only the vectors within those clusters. The IVFx represents the number of samples clusters.

\textbf{PQ (Product Quantization)}: PQ splits the vector into smaller sub-vectors, quantizes each sub-vector to a codebook of possible values. During search, it matches codes in the query to those in the dataset. The PQx represents the number of vector chunks.

\textbf{HNSW (Hierarchical Navigable Small World)}: HNSW builds a multi-level graph where each node represents a vector, and edges connect nodes that are close in the vector space. During search, it starts at the top level and navigates down through the graph, progressively narrowing down the candidates. The HNSWx represents the number of node-connected neighbors for hierarchical small worlds.

The search time statistics for different base methods in All/Typical data are shown in Fig. \ref{fig_6}. We can observe that:

1) With typical scenario data, all the base search algorithms are nearly an order of magnitude faster, while the search accuracy remains at the same level without sacrificing compared to the best Flat results.

2) Among the algorithms, the \textit{HNSW-TSD} is significantly better than IVF, PQ, Flat, etc., with only $3ms$ to complete the search task in typical scenario database. The performance is also influenced by parameters required by the algorithms.

3) \textit{HNSW-TSD} can overcome the problem of involving in the dense common data. In the original database, if the neighbourhood number of HNSW is relatively small (e.g., 16), it may get into the trouble of early stopping and lead to too much deviation of search accuracy. While in typical scenario data, it will be greatly improved. In summary, \textit{HNSW16-TSD} and \textit{HNSW32-TSD} are more suitable for the searching process.

\begin{figure}[htbp]
\centering
\includegraphics[width=2.4in]{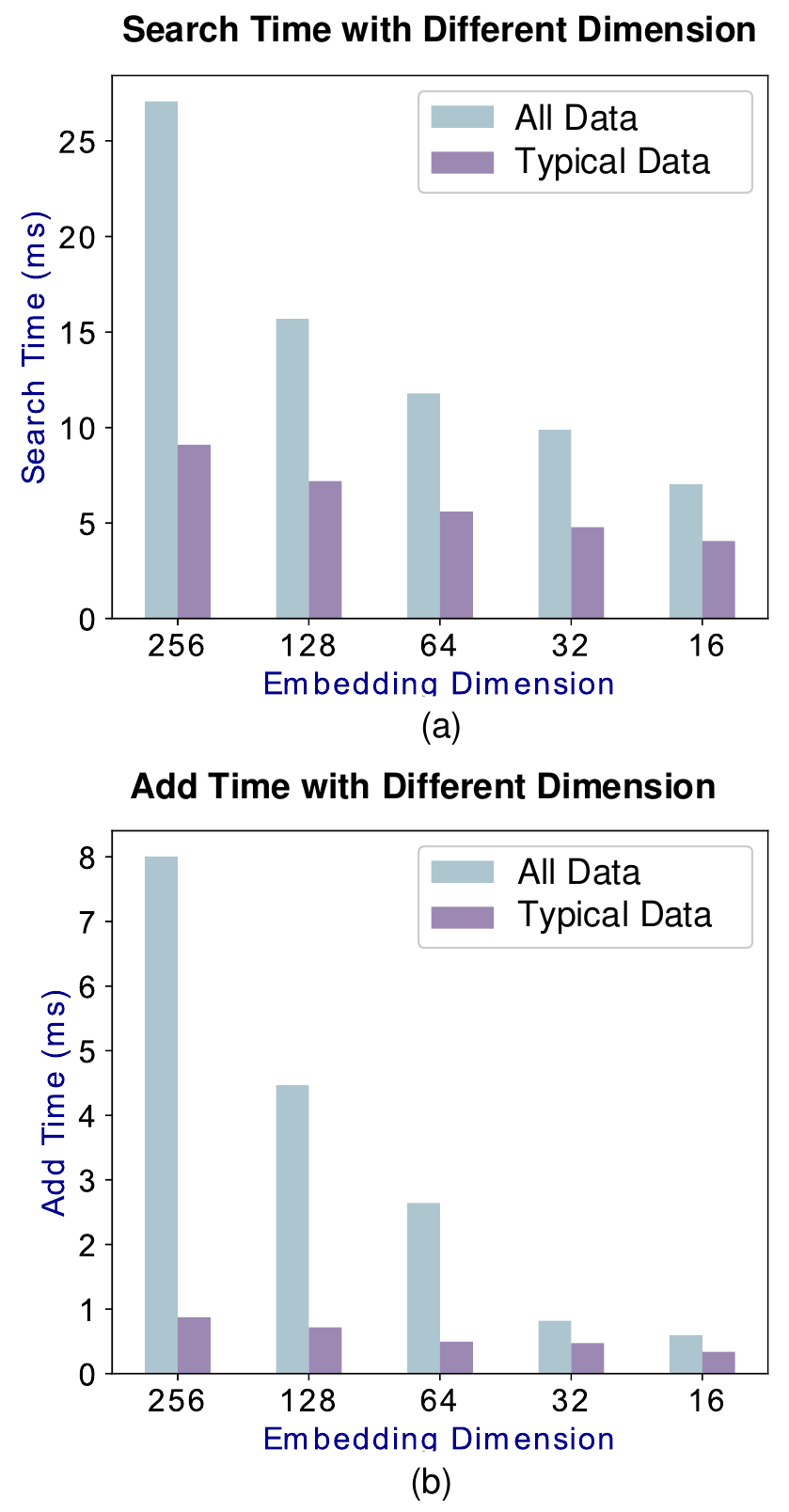}
\caption{Comparison of (a) Search Time and (b) Add Time with different embedding dimensions.}
\label{fig_7}
\end{figure}

When vehicles encounter valuable corner cases, the scenarios will also be added to expand the database. We validate the search/add time with the changes of embedding dimensions, as shown in Fig. \ref{fig_7}. It can be observed that maintaining the embedding dimension at an appropriate value (e.g., 64) can save both search and add time costs, also with the accurate search results shown in Table \ref{tab_1}.

\begin{table*}[htbp]
\centering
% \captionsetup{justification=centering, labelfont={color=blue}}
\caption{Comparison of different parameters ($\alpha$, $\beta$) for \textit{HNSW32-TSD} searching performance.}
\includegraphics[width=6.0in]{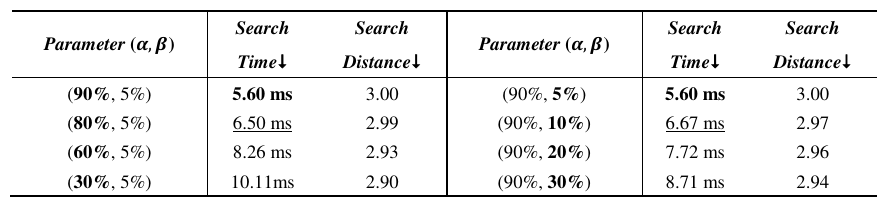}
\label{tab_2}
\end{table*}

Next, as shown in Table \ref{tab_2}, we conduct parameters sensitivity analysis, and compare the search performance of different parameters ($\alpha$, $\beta$) under \textit{HNSW32} base method. As the $\alpha$ parameter decreases from 90\% to 30\%, the number of retained "high-density scenarios" increases, resulting in a significant increase in search time. Due to the relatively rare corner scenarios, the addition of a large number of common scenarios will not improve much search accuracy. As the $\beta$ parameter increases from 5\% to 30\%, the number of sampled low-density scenarios also increases, resulting in a significant increase in search time while little promotion on search scenarios distance. Therefore, ($\alpha$, $\beta$) is selected as (90\%, 5\%) to keep the balance between low-density and high-density scenarios, which achieves better performance on search efficiency and accuracy. For the expansion distance parameter \textit{D}, we generally select a value that is greater than the distance between normal scenarios. The $D$ parameter can also be adjusted based on the scenario distribution and the number of scenarios that researchers tend to expand.

\begin{figure}[htbp]
\centering
\includegraphics[width=2.8in]{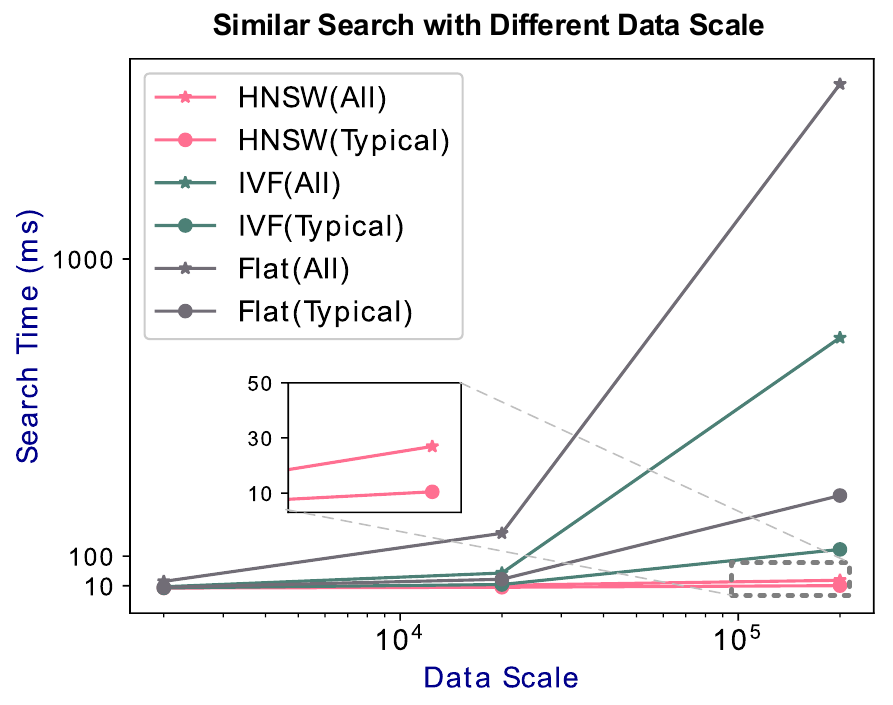}
\caption{Performance comparison for different data scales.}
\label{fig_8}
\end{figure}

Finally, as shown in Fig. \ref{fig_8}, the advantages of the method become more apparent when the size of the data increases. When the number of accumulated scenarios reaches the scale of $10^5$, \textit{HNSW-TSD} can still control the search time to $10 ms$, which is several times or even orders of magnitude ahead of other methods. The trend is also similar for larger amounts of data. While stronger machine performance will alleviate the problem, the proposed method has the ability to improve performance for different data types, data scales and machines. The speed of scenario search process will enhance both the offline simulation systems and online vehicle tasks.

\subsection{The Important Role of Scenario Search for LLM-RAG Systems}
\begin{table*}[htb]
\centering
% \captionsetup{justification=centering, labelfont={color=blue}}
\caption{Comparison of LLM-RAG planning results with different conditions in different scenarios.}
\includegraphics[width=6.3in]{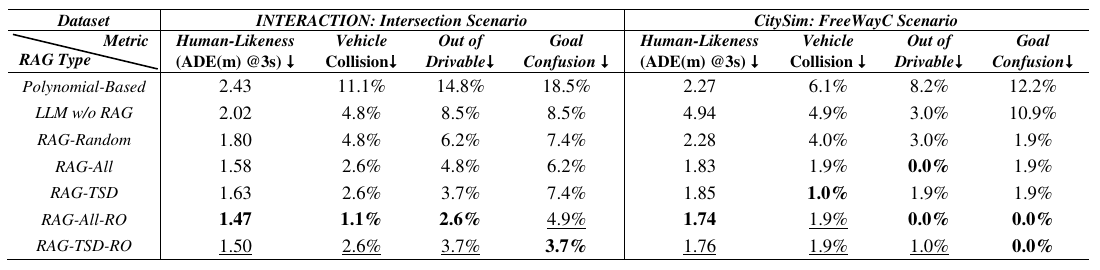}
\label{tab_3}
\end{table*}

In this part, we illustrate the effectiveness for the searched scenario data via typical LLM-RAG trajectory planning application, which is a common way to test the performance of RAG systems \cite{wang2023chatgpt, yuan2024rag}. Here we select \textit{GPT-4o-mini} as the base LLM, which is a lightweight and efficient model. Hundreds of interactive scenarios are selected for testing the task from the two datasets. As shown in Table \ref{tab_3}, we adopt the average displacement error (ADE) between LLM planning trajectory with human driving trajectory in future $3 s$ time horizons to calculate human-likeness, and conduct statistics on the rate of vehicle collisions, out of drivable area, and vehicle goal confusion in long-term planning. We can observe that:

1) Compared to traditional polynomial-based planning method \cite{gasparetto2010optimal}, LLM has certain reasoning ability with well-developed scenarios descriptions and CoTs, which significantly reduces the conditions of driving out of the drivable area and goal confusion. However, there are still relatively large bias in LLM's planning for agents without the augment from retrieved human experiences.

2) After adding randomized scenarios from the scenario database for empirical augmentation (RAG-Random), the metrics are improved. However, random experiences can also lead to LLM hallucination, preventing the system from achieving optimal performance.

\begin{figure}[htbp]
\centering
\includegraphics[width=3.1in]{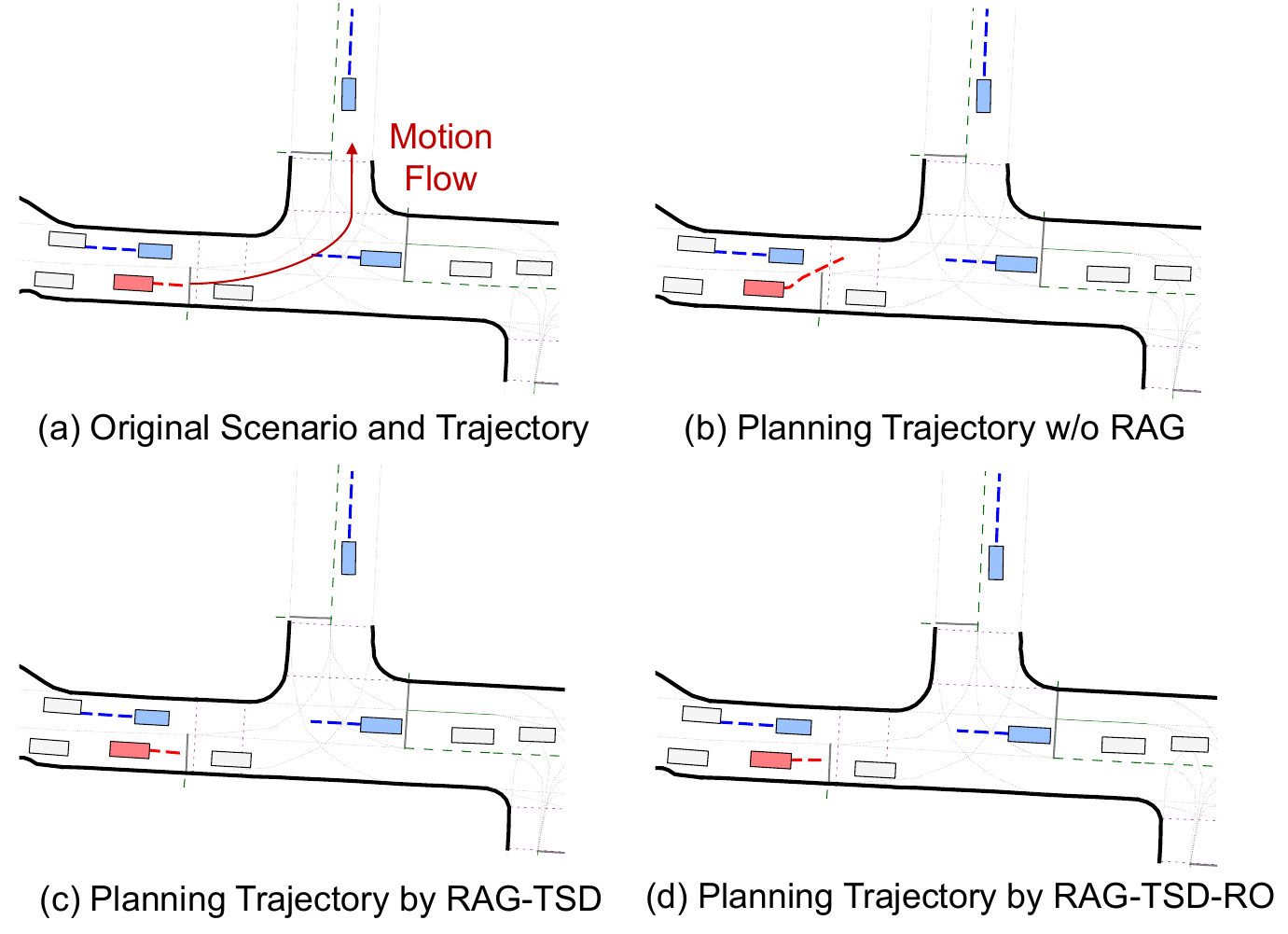}
\caption{Comparison of planning trajectories in \textit{INTERACTION} dataset.}
\label{fig_9}
\end{figure}

\begin{figure}[htbp]
\centering
\includegraphics[width=3.1in]{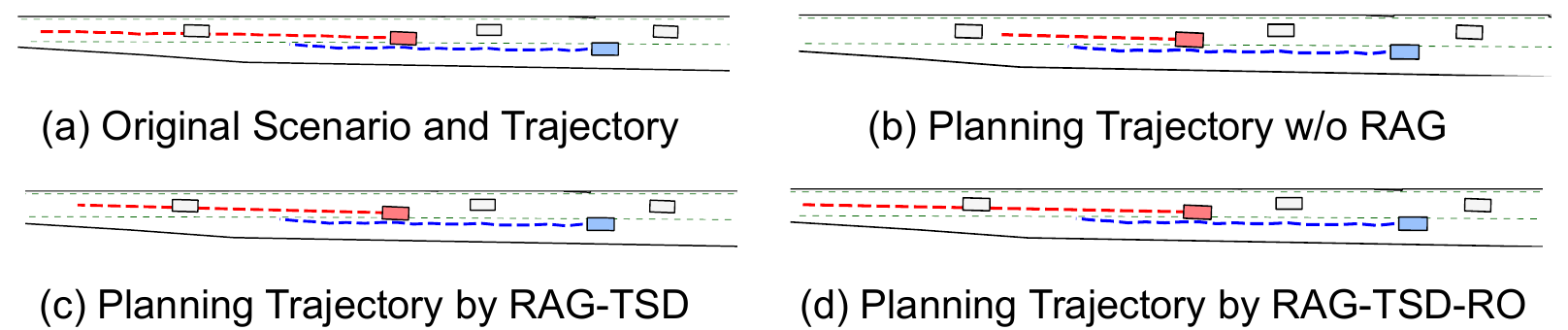}
\caption{Comparison of planning trajectories in \textit{CitySim} dataset.}
\label{fig_10}
\end{figure}

3) When we utilize either the all scenario data or typical scenario data for vector HNSW similarity search, the results of planning are both greatly improved. In particular, RAG in the selected database significantly improves the search speed while still maintaining comparable effectiveness with all data.

4) With Re-Organization (RO) by graph relations, the searched top-$K$ scenarios are further selected to ensure the consistency of relations and intentions, which is reflected in the less deviations and errors in planning tasks. Especially for the goal confusion rate, vehicles are more aware of the destinations and intentions. It suggests that adding knowledge judgments to a few searched data and consuming only tiny amount of time can further improve the results.

Here we select two typical cases to demonstrate the results, as shown in Fig. \ref{fig_9} \ref{fig_10}. In the two datasets, the ego vehicle (depicted in red) is required to navigate through intersection/ramp while managing conflicts and potential interactions with surrounding vehicles. Without the assistance of RAG system, the ego vehicle struggles with understanding the scenario and properly navigating the lanes. In \textit{INTERACTION} dataset, it faces confusion when traveling in the wrong direction, heading straight towards the destination against traffic. In contrast, with the help of retrieval scenarios via RAG-TSD and RAG-TSD-RO, the ego vehicle successfully makes the correct decision to yield the right-of-way as it approaches the stop line. In \textit{CitySim} dataset, the ego exhibits overly cautious behavior, driving too slowly and conservatively. This results in difficulty for rear-end vehicle on ramps to merge and increases the risk of collisions. With RAG, the ego adjusts its velocity to a more human-like level, improving both its flow and interactions.

Despite the relatively more computational time and memory requirements, LLM-RAG has the ability to leverage human experience for reasoning, which helps to minimize inexplicable errors that may occur with traditional planning methods. The human-centric approach allows vehicle to make more intuitive and context-aware decisions, similar to how a human driver would navigate complex scenarios \cite{he2023towards}. For applications, in offline tasks such as simulation and generation, the framework can serve as a crucial reference for integrating and enhancing other planners \cite{sharan2023llm}. Their capacity to understand and generate contextually relevant solutions can provide a more comprehensive and realistic basis for simulations. For real-time decision-making and planning tasks, the potential for LLM-RAG is also growing as the computational power of in-vehicle/cloud platforms gradually increases and the cost of foundation models decreases.

\section{Conclusion}
In this paper, we present Driving-RAG, a driving scenario data-based RAG framework optimized for exceptional speed without compromising accuracy or performance. Specifically, we introduce the scenario embedding model that achieves feature alignment in vector space to provide suitable distinguishable embeddings for scenario data accurately and efficiently. We design the \textit{HNSW-TSD} vector search algorithm, which improves the speed by at least an order of magnitude. The embedding and vector search time costs are compressed to about $30ms$ at the $10^5$ level of data scale via 128G RAM. Further by reorganization with relations extraction, the effectiveness of RAG and our search methods are verified on typical trajectory planning task via LLM. Compared to the general LLM planning method, the performance is significantly improved.

It is worth noting that although our Driving-RAG framework is designed to address challenges in RAG based autonomous driving, it is inherently a generalizable solution that can be applied to a wide range of tasks involving RAG or database systems. In future work, we aim to further explore the collaborative performance achieved by integrating traditional methods with RAG-based approaches and to discuss more applications that can benefit from the framework.


\bibliography{sn-bibliography}% common bib file

\end{document}